\documentclass[a4paper,11pt]{article}

\usepackage[a4paper, left=2.5cm, right=2.5cm, bottom=2cm, top=2cm]{geometry}
\usepackage[unicode, colorlinks=true, linkcolor = black, citecolor=black]{hyperref}
\usepackage[utf8]{inputenc}
\usepackage{graphicx}
\usepackage{mathptmx} 
\usepackage{authblk} 
\usepackage{xcolor}
\usepackage{caption}
\usepackage{amsmath}
\usepackage{amsmath,amsfonts,amssymb}
\usepackage{physics}
\usepackage{adjustbox}
\usepackage{multirow}

\makeatletter
\def\@maketitle{%
  \newpage
  \begin{center}%
  \let \footnote \thanks
    {\large\bfseries \@title \par}%
    \vskip \@affilsep%
    {\large%
        \begin{tabular}[t]{c}%
        \@author\\
        \hyperlink{mailto:\contact}{\contact}
      \end{tabular}\par}%
  \end{center}}
\makeatother

\title{Benefiting from Quantum? A Comparative Study of Q-Seg, Quantum-Inspired Techniques, and U-Net for Crack Segmentation}
\author[1,4]{Akshaya Srinivasan}
\author[1]{Alexander Geng}
\author[2]{Antonio Macaluso}
\author[3,4]{Maximilian Kiefer-Emmanouilidis}
\author[1]{Ali Moghiseh}
\affil[1]{Fraunhofer Institute for Industrial Mathematics (ITWM), Fraunhofer-Platz 1, 67663 Kaiserslautern}
\affil[2]{German Research Center for Artificial Intelligence (DFKI), Campus D 3.2, 66123 Saarbrücken}
\affil[3]{German Research Center for Artificial Intelligence (DFKI), Trippstadter Str. 122, 67663 Kaiserslautern }
\affil[4]{Department of Computer Science and Research Initiative QC-AI, RPTU Kaiserslautern-Landau,
Erwin-Schrödinger-Straße 48, 67663 Kaiserslautern}
\def\contact{}

\begin{document}

\maketitle

\begin{abstract}
\noindent Exploring the potential of quantum hardware for enhancing classical and real-world applications is an ongoing challenge. This study evaluates the performance of quantum and quantum-inspired methods compared to classical models for crack segmentation. Using annotated gray-scale image patches of concrete samples, we benchmark a classical mean Gaussian mixture technique, a quantum-inspired fermion-based method, Q-Seg a quantum annealing-based method, and a U-Net deep learning architecture. Our results indicate that quantum-inspired and quantum methods offer a promising alternative for image segmentation, particularly for complex crack patterns, and could be applied in near-future applications.
\end{abstract}


\section{Introduction} \label{sec:Srinivasan:intro}
Quantum computing has emerged as one of the leading technologies to improve the efficiency and solvability of complex problems. Still, the bridge between fundamental and applied research is very narrow and under construction.
Unsupervised learning emerges as a particularly promising avenue for the adoption of quantum computing in machine learning. Classical algorithms often struggle to efficiently detect patterns in unlabeled data, a common scenario in many practical applications. Recent advancements have showcased the potential of quantum optimization techniques in addressing unsupervised segmentation tasks~\cite{presles2024synthetic, venkatesh2023q}. Furthermore, combining quantum computing with classical methods have led to quantum-inspired (QI) and hybrid methods like hybrid quantum image edge detection~\cite{geng2022hybrid} or quantum transfer learning, which have been used for example for crack detection~\cite{geng2023hybrid}.

In this paper, we want to build on these developments, and furthermore evaluate how quantum effects in quantum and QI methods can be harnessed to advance classical algorithms as well as benchmarking current state-of-the-art approaches. As a use case we have chosen crack-segmentation, a real-world problem, which we consider a tremendously important task to evaluate for example the quality of current roads and infrastructure, see Fig.~\ref{fig:Srinivasan:crack_seg_pipeline} (a). By conducting a systematic comparison between four approaches, where two benefit from quantum, we seek to identify specific areas where non-classical approaches offer advantages. This research not only contributes to the understanding of quantum computing’s practical applications but also guides future developments in algorithm design and implementation within the field.

\section{Segmentation Techniques}\label{sec:Srinivasan:seg_methods}
\begin{figure}[tb]
  \centering
  \includegraphics[width=1\textwidth]{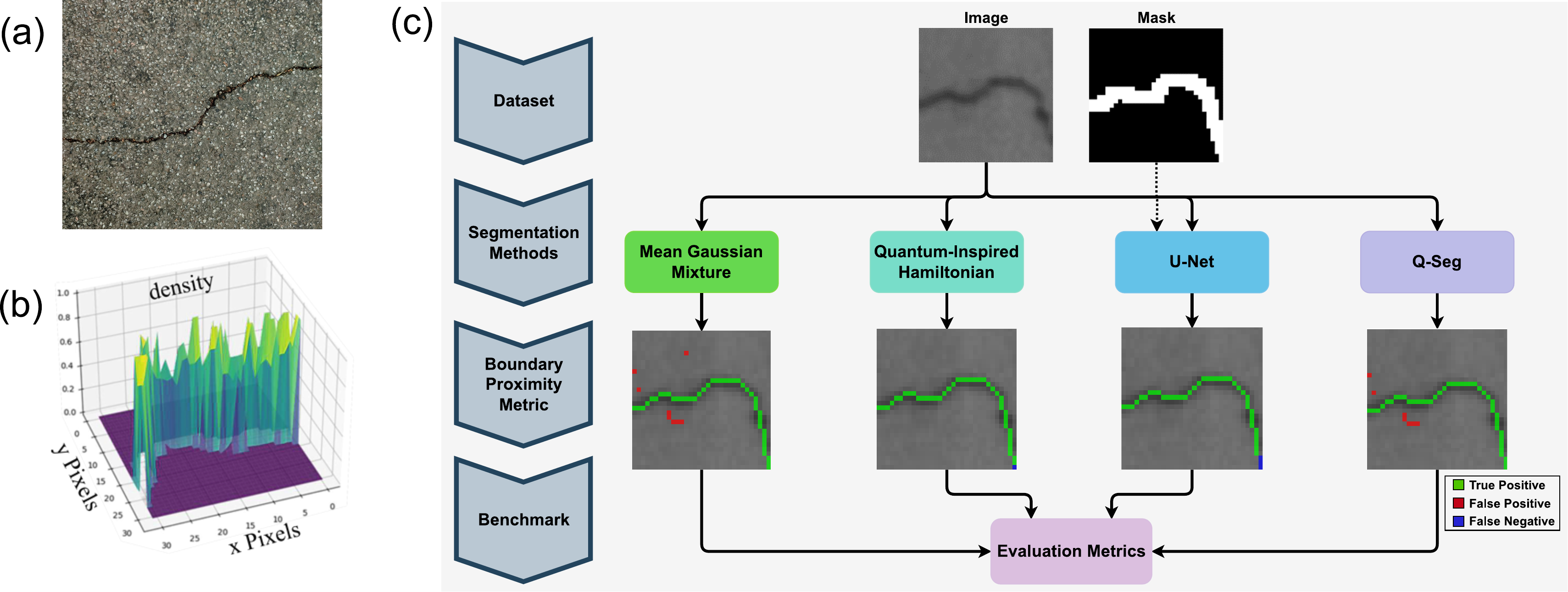}
  \caption{Overview of crack segmentation motivation and methodology: (a) Cracks on roads illustrating real-world infrastructure challenges, (b) Results from the QI approach, accurately identifying crack locations using localized states tied to negative eigenvalues, and (c) Comparative pipeline of crack segmentation methods. }
  \label{fig:Srinivasan:crack_seg_pipeline}
\end{figure}

This section examines four methodologies for segmenting concrete cracks: Mean Gaussian Mixture (MGM), QI Hamiltonian, Q-Seg, and U-Net. Using a dataset of $32 \times 32$ pixel images annotated with ground truth crack locations, each method processes input images to generate segmentation masks that delineate detected cracks. The approaches differ in complexity and computational demands, reflecting advancements in classical and quantum techniques. Figure~\ref{fig:Srinivasan:crack_seg_pipeline} (c) provides a comparative overview of these workflows.

\subsection{Mean Gaussian Mixture} \label{subsec:Srinivasan:threshold}

The Gaussian mixture model is a fundamental image segmentation technique known for its simplicity and efficiency, especially when objects of interest, like pores, have distinct intensity levels. This classical method is computationally inexpensive and versatile, making it ideal for preliminary segmentation tasks. In our study, we adapt Otsu Thresholding \cite{otsu1975threshold} to segment cracks in concrete images. Otsu's method determines the optimal threshold that maximizes between-class variance, effectively separating the foreground (cracks) from the background.
To ensure consistent intensity across all samples, each image is normalized to a range of $[0, 255]$, addressing ambient lighting variations. Otsu method is applied to 30 images, calculating the optimal threshold for each. For consistency in our comparative analysis, we use the mean threshold across these images as a global threshold, allowing us to benchmark different segmentation methods. This approach balances individual image optimization and comparative consistency across the dataset.

\subsection{Quantum-Inspired Hamiltonian} \label{subsec:Srinivasan: hamiltonian}
Due to the rise of quantum computing also QI-techniques have become more prominent in image processing \cite{Singh2024}. In the original context, QI refers to the idea to evaluate classically how quantum effects like superposition, entanglement or wave function collapse (measurements) may change an algorithm of interest \cite{moore1995quantum, huynh2023quantuminspiredmachinelearningsurvey}, and in the best case how to benefit from it. Simulating general many-body quantum systems and circuits becomes exponentially difficult as the number of particles or qubits increases. However, many problems can be reduced to polynomial complexity. For example, single-particle Hamiltonians allow each particle to be evaluated separately, with the combined dynamics described as presented here by Fermi-Dirac  statistics \cite{IngoPeschel_2004}. We refer to \cite{Palaiodimopoulos2024} and \cite{Dutta2021} for a deeper discussion of the underlying physical effect, fitting in the context of this work. 
In this paper, we show in a proof-of-concept that the single-particle effect of Anderson Localization (AL) \cite{abrahams201050, Anderson1958} can be efficiently used for image- and especially crack-segmentation. Initially used to explain electron behavior in disordered lattices, this model introduces randomness into the potential energy landscape, leading to the localization of wave functions, see Fig.\ref{fig:Srinivasan:crack_seg_pipeline} (b). Effects of AL and disorder have been found suitable for image and signal processing tasks such as image representation and denoising \cite{Dutta2021}, augmentation \cite{Palaiodimopoulos2024, palaiodimopoulos2024disquinvestigatingimpactdisorder} and signal transfer in optical fibre \cite{Mafi2021}. Embedding an image in a Hamiltonian yield a simple matrix form $N\times N$, where $N$ correspond to lattice sites or qubits. To formulate our Hamiltonian matrix, we slightly adjust the adaptive signal decomposition presented in \cite{Dutta2021}. The key QI idea is that the embedding of the images themselves will self-induce AL due to their disordered landscape, i.e. rough surfaces and cracks which seem like random disordered potentials. Thus, the eigenstates of the Hamiltonian will be close to a unit vector  (hot encoded one), only in the strongly disordered areas, and rather extended in areas of weaker disorder. In our study, we embed the image on a 2D lattice. Here it is known that the localization length scales exponentially with disorder strength as well as energy. Thus leading to strong dependence on disorder effects, as particles will mostly localize in areas of the cracks and holes. The embedding works as follows.
First, the $m \times n$ images are flattened $m \cdot n$, such that a pixel value at $A_{ij}\rightarrow a_l$ with $l = j + ni$. The corresponding single-particle Hamiltonian of $N \times N$ size where $N=m\cdot n$ reads
\vspace{-5pt}
\begin{equation}\label{eq:Srinivasan:ham}
\small
\begin{aligned}
      H_{i,j} =
            \begin{cases} 
            a_i & \text{if } i = j \\
            G(a_i, a_j) & \text{if } |i-j| =1 \text{ and }\\
            & \quad i,j\mod n \neq 0 \\
            G(a_i, a_j) & \text{if } |i - j| = n \\
            0 & \text{otherwise}
            \end{cases}&, \quad  & G(a_i, a_j) = \exp\left( - \frac{(a_i - a_j)^2}{2\sigma^2} \right)
\end{aligned}
\end{equation}

The $G(a_i, a_j)$ is the Gaussian difference only for nearest-neighboring pixels and $\sigma^2$ is the Gaussian variance. The diagonal elements of the Hamiltonian matrix $a_{i}$  correspond to the pixel values (potentials), while the off-diagonal $G(a_i,a_j)$ elements represent the Gaussian weights between nearest-neighboring pixels (kinetic terms).
The Hamiltonian in its diagonal form can already be considered as a thresholding technique, however, inferior to the MGM explained in Sec.~\ref{subsec:Srinivasan:threshold}. Only due to the kinetic terms we will get extended states which do not contribute significantly to the density of the crack or the whole picture at all. However, we have to be careful to construct kinetic terms for nearest neighbours, as otherwise we might generate extended states again \cite{palaio2023}.
Furthermore, we have found the Gaussian distance better than constant kinetic terms in \cite{Dutta2021}. The final mask shows the elementwise summation of magnitudes of all eigenstates (localized particles), tied only to negative eigenvalues. The sum of all negative eigenvalues corresponds to the many-body ground state energy and thus is the minimal energy of the system. This method shows surprising good results and efficiently finds the crack, see Fig.\ref{fig:Srinivasan:crack_seg_pipeline} (b).

\subsection{Q-Seg: Unsupervised Quantum Algorithm} \label{subsec:Srinivasan:qseg}

Q-Seg is an innovative image segmentation method that utilizes quantum annealing \cite{RevModPhys.80.1061, farhi2000quantumcomputationadiabaticevolution}. Initially tested for Earth observation images \cite{venkatesh2023q}, Q-Seg adapts to detect cracks in concrete by efficiently solving the Maxcut problem using a D-Wave quantum annealer. 

The segmentation procedure begins by converting the input image into a lattice graph where each pixel is a node, preserving spatial connectivity. Edges are weighted based on pixel similarity, calculated as squared differences, in our case to enhance contrast to gray-scale crack images. The segmentation task becomes a graph cut problem, aiming to find a maximum cut that best partitions the vertices based on edge weights. To overcome the computational challenges of finding maximum cuts, Q-Seg reformulates the problem into a Quadratic Unconstrained Binary Optimization (QUBO) formulation, suitable for quantum annealing \cite{neven2008imagerecognitionadiabaticquantum}.
The QUBO problem is mapped onto the \textit{Pegasus} architecture \cite{dattani2019pegasussecondconnectivitygraph} of the D-Wave quantum annealer, where the system starts in a superposition of all possible states and gradually evolves toward the lowest energy state that represents the optimal solution.  The D-Wave quantum annealer iteratively adjusts system parameters and annealing cycles.  This iterative adjustment increases the probability of reaching the global minimum.

The final result of the quantum annealing process is a binary string corresponding to the segmented image, providing a direct solution to the image segmentation problem. This unsupervised segmentation approach proved effective beyond its original Earth observation application in adapted scenarios such as crack detection in concrete structures, demonstrating Q-Seg's versatility and robustness in various image segmentation tasks.

\subsection{U-Net} \label{subsec:unet}
U-Net is a deep-learning architecture designed for biomedical image segmentation \cite{ronneberger2015u}, renowned for its performance in tasks with limited annotated data. Its versatility extends to various applications, including medical imaging, satellite imagery, and material defect detection. This study focuses on utilizing U-Net for crack detection in concrete, leveraging its strength in producing detailed segmentation masks.

The U-Net architecture features a U-shape consisting of an encoder and a decoder. The encoder reduces the spatial dimensions of the input image using convolutional and max-pooling layers, creating a lower-resolution representation. The decoder upsamples the image, restoring lost spatial dimensions. A notable advantage of U-Net is its skip connections between encoder and decoder layers, which allow access to high-resolution feature maps, improving segmentation accuracy. For binary segmentation tasks like crack detection, a sigmoid activation function classifies each pixel as a crack or background.

In this study, U-Net architecture is modified to handle $32 \times 32$ pixel crack images, trained on a dataset of 456 labeled patches. The model, with approximately 21.7 million trainable parameters, generates a binary mask indicating crack presence. The training utilized a batch size of 16 over 50 epochs, completed in about 13 minutes on a local machine with an Intel Core i7 CPU and 16GB of RAM. U-Net's ability to capture detailed features makes it a valuable tool for precise and reliable crack segmentation.

\section{Dataset and Metrics for Segmentation Analysis} \label{sec:Srinivasan:analysis}
\begin{figure}[tb]
    \centering
    \includegraphics[width=0.85\textwidth]{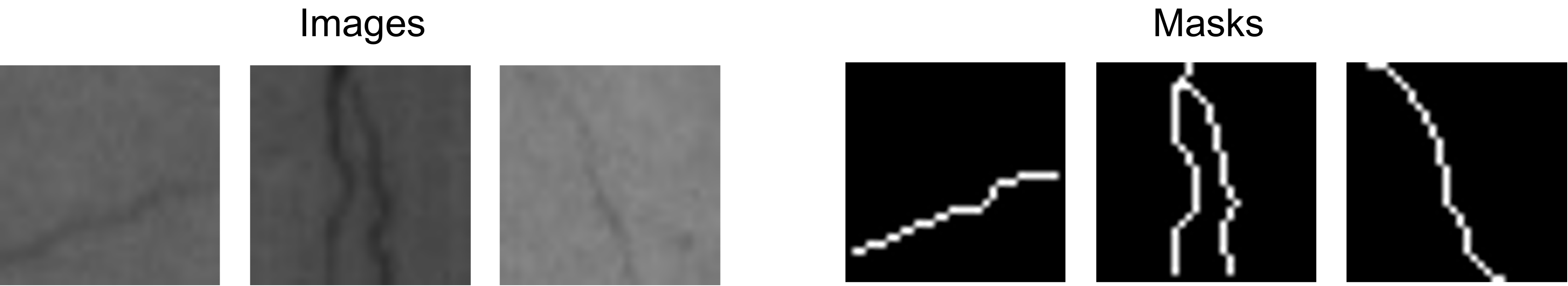}
    \caption{Sample images of cracks with corresponding masks.}
    \label{fig:Srinivasan:cracks}
\end{figure}

This study utilizes grayscale images of concrete for crack segmentation using four different methods. The original images measure approximately $16,000 \times 32,000$ pixels, with cracks only $1-3$ pixels wide, making detection challenging for the human eye and machine learning algorithms. To accommodate the limitations of the D-Wave quantum annealer, including the restricted number of qubits and limited runtime, we divide the images into smaller $32 \times 32$ pixel patches. Our complete dataset consists of 456 manually annotated patches, split into $70\%$ for training and $30\%$ for validation of the U-Net model. We evaluate the performance of the segmentation methods—MGM, QI Hamiltonian, Q-Seg, and U-Net on an unseen test dataset of $30$ patches. Figure \ref{fig:Srinivasan:cracks}
presents example patches with manually annotated masks highlighting the cracks, which are used as ground truth for performance comparison.
\vspace{-10pt}
\subsection{Evaluation Metrics} \label{subsec:Srinivasan:metrics}
To evaluate the segmentation methods, we use the confusion matrix, F1 score, and Intersection over Union (IoU). The confusion matrix provides four key metrics: True Positives (TP), False Positives (FP), False Negatives (FN), and True Negatives (TN), which assess the accuracy of crack predictions. The F1 score combines precision and recall, while IoU measures the overlap between predicted and ground truth masks, providing a comprehensive evaluation of segmentation performance.
\vspace{-5pt}
\begin{equation}\label{eq:Srinivasan:F1andIou}
\small
\begin{aligned}
     F1 ~ Score = 2 \times \frac{\text{Precision} \times \text{Recall}}{\text{Precision} + \text{Recall}} & \quad & IoU = \frac{TP}{TP + FP + FN}.
\end{aligned}
\end{equation}

\subsection{Boundary Proximity Metric} \label{subsubsec:Srinivasan:bpm}
In traditional segmentation tasks, evaluation metrics such as the confusion matrix 
may not adequately reflect performance when slight deviations in boundary prediction occur. In crack segmentation tasks, where cracks are typically thin structures with irregular boundaries, these minor deviations should be tolerated to some extent. The Boundary Proximity Metric (BPM) addresses this by adjusting the boundary around the cracks in both the predicted segmentation \textbf{$I_P$} and the ground truth \textbf{$I_{GT}$}, allowing for more lenient evaluation in cases of minor misalignment. The process starts by skeletonizing both the ground truth \textbf{$S(I_{GT})$} and predicted segmentation results \textbf{$S(I_P)$}. Skeletonization reduces each crack to its core structure, which helps in focusing only on the most critical regions. After skeletonization, both the ground truth and the predicted results are dilated using flat disk structuring element \textbf{$B_r$} by a radius of \textbf{$r$ }pixels. This dilation adjusts the boundary, expanding it to account for small deviations. Any predicted crack pixels that were previously identified as false positives or false negatives but fall within this dilated boundary (i.e., within $r$ pixels of the ground truth) are then reassigned as true positives. This re-calibration of TP, TN, FN and FP are mathematically formulated as follows
\begin{equation}
\label{eq:Srinivasan:bpm}
\small
\begin{aligned}
    I_{\widetilde{TP}} &= \left[S(I_{GT}) \oplus B_r\right] \cap S(I_P),&
    I_{\widetilde{FP}} &= \left[\left[S(I_{GT}) \oplus B_r\right] \cap S(I_P)\right] - S(I_P), \\
    I_{\widetilde{FN}} &= \left[\left[S(I_{P}) \oplus B_r\right] \cap S(I_{GT})\right] - S(I_{GT}), & 
     I_{\widetilde{TN}} &= I_{ones} - \sum \left(I_{\widetilde{TP}} + I_{\widetilde{FP}} + I_{\widetilde{FN}}\right),
\end{aligned}
\end{equation}
where \textbf{$I_{ones}$} is $n \times n$ matrix with all entries equal to 1 and $n$ is the size of the crack image.
The new counts for true positives $\widetilde{TP}$, false positives $\widetilde{FP}$, false negatives $\widetilde{FN}$, and true negatives $\widetilde{TN}$ are calculated by applying Eq.~\ref{eq:Srinivasan:bpm} and
\vspace{-2pt}
\begin{equation}\label{eq:Srinivasan:bpm-counts}
\small
\begin{aligned}
     \widetilde{X} = \norm{I_{\widetilde{X}}}_1 = \sum_{i}\sum_{j} I_{\widetilde{x}} (i, j) & \quad \text{where} ~ & \widetilde{X} \in \left[\widetilde{TP}, \widetilde{FP}, \widetilde{FN}, \widetilde{TN}\right].
\end{aligned}
\end{equation}
Using this boundary proximity metric makes the evaluation more forgiving towards minor misalignment that would otherwise result in a higher count of false positives and false negatives. This approach is especially beneficial in crack segmentation, where small discrepancies in boundary prediction are often unavoidable due to the irregular shapes of cracks.

\section{Results and Discussion} \label{sec:Srinivasan:results}

\begin{figure}[tb]
    \centering
    \includegraphics[width=1\textwidth]{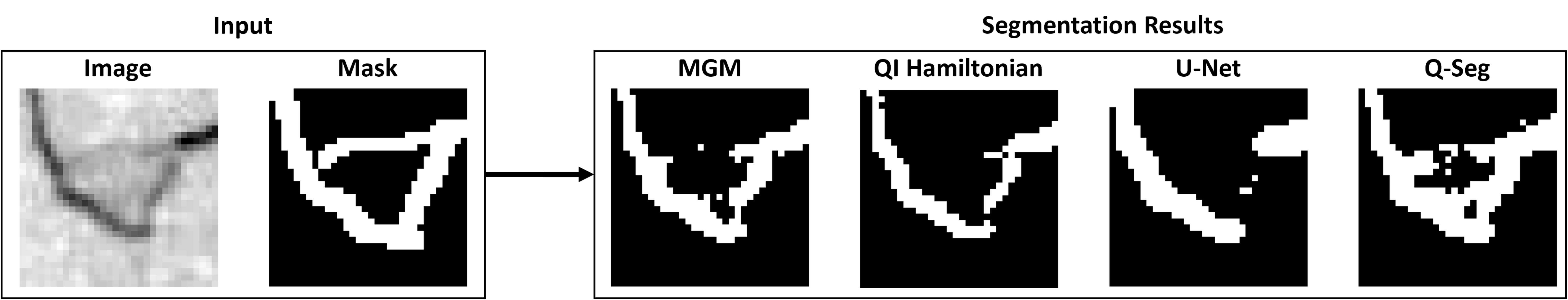}
    \caption{Crack segmentation results from four different techniques: MGM, the QI Hamiltonian method, U-Net, and Q-Seg.}
    \label{fig:Srinivasan:crack-res}
\end{figure}

\begin{figure}
    \centering
    \includegraphics[width=1\linewidth]{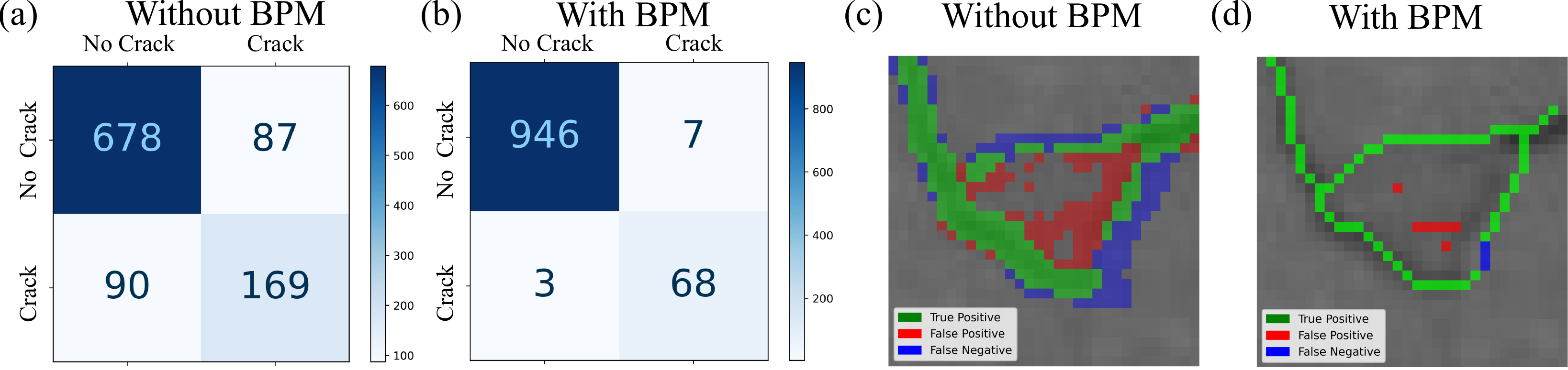}
     \caption{Visual comparison of segmentation results, including the standard confusion matrix (a), the confusion matrix post-BPM application (b), and an overlay of predicted crack masks against actual cracks before (c) and after BPM (d).}
     \label{fig:Srinivasan:res-bpm}
 \end{figure}

We have benchmarked MGM, QI Hamiltonian, Q-Seg, and U-Net using standard evaluation metrics and prediction time for segmenting 30 images. Additionally, we employ the BPM to refine the evaluation by considering slight deviations in the predicted crack boundaries compared to the ground truth.
Each segmentation method shows distinct results in detecting cracks, as illustrated in Figure~\ref{fig:Srinivasan:crack-res}. This figure underscores the strengths and limitations of each approach in capturing fine details and improving prediction accuracy.
Furthermore, Figure~\ref{fig:Srinivasan:res-bpm} demonstrates the visual comparison of the segmentation results, showing both the standard confusion matrix and the one after applying BPM. An overlay diagram illustrates the alignment between the predicted crack masks and the actual cracks. This comparison emphasizes the impact of BPM in improving segmentation accuracy, particularly in challenging cases where the cracks are faint or unclear.
Table~\ref{table:Srinivasan:metrics} provides a detailed comparison of the segmentation methods, both with and without BPM adjustments, highlighting their effectiveness in crack detection. The table includes average F1 scores and IoU values for each segmentation technique, allowing for a comprehensive performance assessment. We also present the corresponding prediction times to provide insight into the computational efficiency of each segmentation task.
\begin{table}[tb]
    \caption{Performance comparison of four crack segmentation techniques with and without boundary proximity metric (BPM) using standard evaluation metrics}
    \centering
    \small
    \begin{adjustbox}{max width=\textwidth}
    \begin{tabular}{|l|c|c|c |c|c|}
        \hline
        \multirow{2}{*}{Segmentation Methods} & \multicolumn{2}{c|}{Metrics without BPM} & \multicolumn{2}{c|}{Metrics with BPM} & \multirow{2}{*}{Prediction Time (s)} \\
        \cline{2-5}
        & Avg IoU & Avg F1 Score & Avg IoU & Avg F1 Score & \\
        \hline
        MGM           &  0.5783 $\pm$ 0.1611 & 0.7197 $\pm$ 0.1449 & 0.7454 $\pm$ 0.1836

  & 0.8439 $\pm$ 0.1781

 &0.032 $\pm$ 0.007 \\
       \hline
        QI Hamiltonian & 0.6218 $\pm$ 0.178 & 0.7478 $\pm$ 0.1766 & 0.9447 $\pm$ 0.1241
 & 0.9693 $\pm$ 0.1016
 & 156 $\pm$ 10\\
        \hline
        U-Net                  & 0.6159 $\pm$ 0.1440 & 0.7522 $\pm$ 0.1145 & 0.8945 $\pm$ 0.1834
 & 0.9395 $\pm$ 0.1697
 & 2.292 \\
        \hline
        Q-Seg                  & 0.5728 $\pm$ 0.1687 & 0.7079 $\pm$ 0.1735 & 0.8014 $\pm$ 0.1431
  & 0.8753 $\pm$ 0.1357
 & 2.277 $\pm$ 0.250 \\
        \hline
    \end{tabular}
    \end{adjustbox}
   \label{table:Srinivasan:metrics}
\end{table}
From Table~\ref{table:Srinivasan:metrics}, we observe that the QI Hamiltonian method delivers the best overall segmentation performance, with the highest average IoU and F1 score using BPM. Without BPM, it performs within the same error bounds as U-Net, showcasing its robustness. However, its prediction time is significantly longer, which limits its efficiency for real-time applications, especially on larger datasets. U-Net, while slightly behind the QI Hamiltonian method in segmentation accuracy with BPM, is still highly competitive, especially without BPM. However, it demands considerable computational resources, and its training time of 13 minutes for 456 $(32 \times 32)$ samples is not included in the prediction time. Q-Seg has a prediction time similar to U-Net, which includes only the Quantum Processing Unit (QPU) access and qubit embedding time. Though it is not as accurate as QI Hamiltonian or U-Net, presents a competitive alternative with balanced performance and does not require labeled data for training, making it practical for scenarios where training data is limited. The MGM model performs comparably to Q-Seg but slightly worse with BPM. However, it is the fastest method, avoiding any training phase like U-Net. Despite its speed, this method lacks segmentation accuracy and is unlikely to perform well with a single mean threshold on larger, more complex datasets. Overall, the outcomes suggest that while U-Net and the Hamiltonian method offer the highest accuracy, Q-Seg provides a balanced alternative with moderate performance and no training requirements.

Future work could focus on testing these approaches on larger datasets to assess their effectiveness in realistic scenarios. Optimizing the Hamiltonian method for GPU parallel processing could yield a 20x speedup \cite{cupy_learningsys2017} and exploring quantum simulations using ultra-cold gas setups \cite{PhysRevResearch.6.033039,bakr2009quantum}. Additionally, exploring Q-Seg on gate-based quantum computing \cite{venkatesh2024qubitefficientvariationalquantumalgorithms} and identifying other challenging domains for annotated data can enhance the application of quantum methods.

\section{Acknowledgements}
We'd like to thank the Quantum Initiative Rhineland-Palatinate (QUIP) for their support. This work was also partially funded by the Research Initiative 'Quantum Computing for Artificial Intelligence' (QC-AI) and the Federal Ministry for Economic Affairs and Climate Action through the EniQmA project (funding number 01MQ22007A).


\end{document}